\documentclass[nohyperref]{article}

\usepackage{microtype}
\usepackage{graphicx}
\usepackage{subfigure}
\usepackage{booktabs} 
\usepackage{pifont}
\usepackage{amsmath}
\usepackage{amssymb}
\usepackage{mathtools}
\usepackage{amsthm}
\usepackage[dvipsnames]{xcolor}
\usepackage{enumitem}
\usepackage{multirow}

\usepackage{hyperref}
\usepackage{enumitem}

\newcommand{\cmark}{\ding{51}}%
\newcommand{\xmark}{\ding{55}}%
\newcommand{\green}[1]{\textcolor{ForestGreen}{#1}}
\newcommand{\yellow}[1]{\textcolor{olive}{#1}}
\newcommand{\red}[1]{\textcolor{BrickRed}{#1}}
\usepackage[accepted]{icml2022_pods}

\usepackage[capitalize,noabbrev]{cleveref}

\theoremstyle{plain}

\theoremstyle{definition}

\theoremstyle{remark}

\usepackage[textsize=tiny]{todonotes}

\icmltitlerunning{A Meta-Analysis of Distributionally-Robust Models}

\begin{document}

\twocolumn[
\icmltitle{A Meta-Analysis of Distributionally Robust Models}



\icmlsetsymbol{equal}{*}

\begin{icmlauthorlist}
\icmlauthor{Benjamin Feuer}{sch}
\icmlauthor{Ameya Joshi}{sch}
\icmlauthor{Chinmay Hegde}{sch}
\end{icmlauthorlist}
\icmlaffiliation{sch}{New York University, New York, USA}

\icmlcorrespondingauthor{Benjamin Feuer}{bf996@nyu.edu}
\icmlcorrespondingauthor{Ameya Joshi}{ameya.joshi@nyu.edu}
\icmlcorrespondingauthor{Chinmay Hegde}{chinmay.h@nyu.edu}

\icmlkeywords{Machine Learning, Deep Learning, Distributional robustness, Computer Vision, Vision-Language Models, ICML}

\vskip 0.3in
]



\printAffiliationsAndNotice{\icmlEqualContribution} 

\begin{abstract}
State-of-the-art image classifiers trained on massive datasets (such as ImageNet) have been shown to be vulnerable to a range of both intentional and incidental distribution shifts. On the other hand, several recent classifiers with favorable out-of-distribution (OOD) robustness properties have emerged, achieving high accuracy on their target tasks while maintaining their in-distribution accuracy on challenging benchmarks. We present a meta-analysis on a wide range of publicly released models, most of which have been published over the last twelve months. Through this meta-analysis, we empirically identify four main commonalities for all the best-performing OOD-robust models, all of which illuminate the considerable promise of vision-language pre-training.
\end{abstract}

\section{Introduction}
\label{sec:intro}

This meta-analysis catalogs the out-of-distribution (OOD) robustness of a variety of image classifiers, with emphasis on models that have been publicly released over the last several months. The goal of our meta-analysis is to address the following question:

\begin{quotation}
\emph{What makes an OOD-robust classifier?}
\end{quotation}

\begin{figure*}[t]
  \centering
  \includegraphics[width=0.8\linewidth]
                  {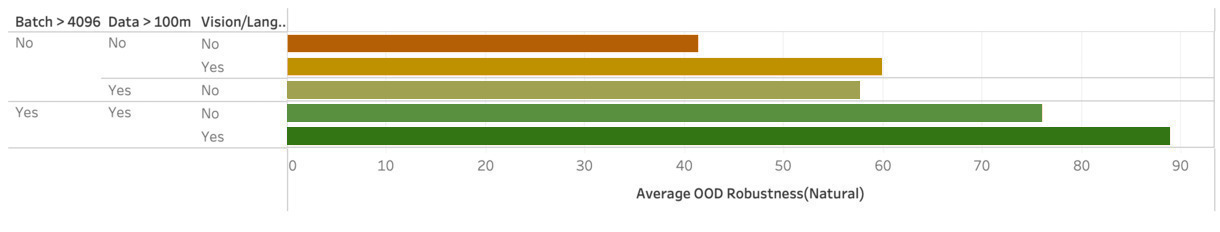}
  \caption{\textbf{Large vision-language models are consistently more OOD-robust under natural distribution shifts. }\sl Robustness of several ImageNet classifiers discussed in our results. Robustness correlates with vision-language contrastive pretraining with high-capacity image backbones trained on massive datasets; robustness reliably fails to develop when any of these properties are absent. Large batch sizes also appear to play a role, but the effect is less consistent.}
  \label{fig:acc_results}
\end{figure*}

By parsing published results (as well as generating our own test results on published models), we uncover several interesting commonalities between the best OOD-robust models. See Figure~\ref{fig:acc_results} for a summary.

\subsection{Setup}

\paragraph{Defining OOD-robustness.} The scope of our analysis corresponds to models capable of inference on the 1000-class ImageNet validation set~\cite{5206848}. 

There are numerous ways to formally define robustness to out-of-distribution shifts. Some emphasize the difficulty of the novel dataset as a component of the loss~\cite{Ben-David2010,harderordifferent}, and others~\cite{Recht2019DoIC} focus on pure gaps in accuracy. This topic has also been studied under related sub-fields such as domain generalization~\cite{Wang2021GeneralizingTU}, and has found applications in several areas including robotics~\cite{Krber2021ComparingPS}.

In our study, for simplicity, we define an ``OOD-robust classifier'' as one that fulfills the following properties:
\begin{enumerate}[noitemsep, parsep=0pt]
\item Attains a high-accuracy regime on the target task, and
\item Maintains $\ge 90\%$ of its in-distribution accuracy under distribution shift using, either zero-shot classification or linear probing (without feature re-training).
\end{enumerate}
In our study, we fix inference on ImageNet as the in-distribution task, and we use a suite of distribution shift datasets designed for ImageNet as OOD benchmarks. 

\paragraph{Metrics of interest.} For inference on the ImageNet validation dataset, we define three broad regimes: the \emph{low} regime covers models with accuracy $\le 50\%$, the \emph{medium} regime covers models with $50-75\%$ accuracy, and the \emph{high} accuracy regime covers models with accuracy $\ge 75\%$. Note that Point \#1 above in our definition of OOD-robustness above automatically excludes the possibility of a low- or medium-accuracy model being labeled OOD-robust. In any case, as anticipated we found that no low- or medium-accuracy classifier achieves out-of-distribution accuracy $\ge 90\%$. 



We consider natural distribution shifts to be those generated by processes such as occlusion, weather, a physical transformation of the object, or varied representations of an object (such as sketches, clip-art, or silhouettes). Similarly, we consider artificial distribution shifts to be those generated by a machine or a mechanical process such as jpeg compression artifacts, blur, noise, and algorithmically generated stylization effects. We do not consider other distribution shifts such as adversarial robustness in this paper, leaving this aspect to future work.

\subsection{Core Findings}

Our most important findings are shown in Figure~\ref{fig:acc_results}. We see that OOD-robustness is undergridded by multiple design and implementation decisions, including both familiar features (massive dataset sizes, training time, model capacity) but also more recent approaches such as large-batch contrastive pretraining with a vision-language training objective.

As we discuss in depth in Sec.~\ref{subsec:model_archs}, we find that four key design decisions reliably co-occur in OOD-robust models, and that the absence of any of these four reliably causes the model to fail to achieve OOD-robustness. We further show that one additional design decision, batch size, has a less predictable (but persistent) effect.

\section{Related Work}
\label{sec:related_work}

\subsection{Natural Distribution Shifts}

Concerns over methodology and generalization in ImageNet-trained classifiers have led to the creation of new datasets as supplements to standard ImageNet validation~\cite{Beyer2020AreWD} ~\cite{Recht2019DoIC}. 
Recent large-scale studies~\cite{Taori2020MeasuringRT,Miller2021AccuracyOT,DBLP:journals/corr/abs-1905-10498,harderordifferent} have suggested that not only do standard ImageNet classifiers generalize less well than we might expect, but they do so in patterned and consistent ways. Our goal in this meta-review is to probe similar patterns in better-performing (distributionally robust) classifiers.

\subsection{Artificial Distribution Shifts}

Training classifiers that are resilient to adversarial perturbations have been shown to produce models less susceptible to spurious features~\cite{Ilyas2019AdversarialEA}. Various forms of artificial image perturbations have also been used for data augmentation ~\cite{Cubuk2020RandaugmentPA}. We note here that OOD-robust models exhibit properties that are distinct from adversarial robustness, in that they attempt to perform well against real-world inputs which are corrupted or otherwise of low quality~\cite{Hendrycks2019BenchmarkingNN}.

\subsection{Vision-Language Contrastive Pretraining}

Contrastive pretraining of visual-language models has opened up exciting new possibilities for designing and scaling robust models. Although precursors of these models have been around for a while~\cite{NIPS2013_7cce53cf,Li2017LearningVN}, several success stories have emerged in the last twelve months. OpenAI's CLIP~\cite{Radford2021LearningTV} was the first vision-language model to reach ImageNet accuracies comparable to SOTA supervised models. Moreover, CLIP also exhibits a surprising degree of OOD-robustness, unlike standard CNN based architectures. Subsequent models trained in a similar fashion, including ALIGN~\cite{Jia2021ScalingUV}, and BASIC~\cite{Pham2021CombinedSF}, have also been able to achieve CLIP-level performance. The cumulative effect has lead some researchers to refer to them as 'more human'~\cite{geirhos2021partial}. Unfortunately, the training details of many of these models, and the associated datasets, remain private.

More recent work has experimented with CLIP-like objectives with smaller public datasets, both using the original CLIP objective and variations thereof. Meta's SLIP models~\cite{mu2021slip} employ a multi-task learning framework for combining self-supervised learning and CLIP pre-training. CLIP, however, remains the only publicly available model which achieves high accuracy on its target task, making apples-to-apples comparisons difficult.


\section{Experimental Covariates}
\label{sec:properties}

Inspired by previous work~\cite{Miller2021AccuracyOT}, we selected a representative sample of classifiers for diversity with respect to properties we originally suspected might play a role in model robustness: number of parameters, noisy label pretraining, vision-language contrastive pretraining, other forms of contrastive pretraining, quantity of data, batch size, and choice of vision backbone.

We group our findings into models for which both artificial and natural distribution shifts are considered, and models where only natural distribution shifts are considered. We do this because some of the models we used for comparison are not publicly available and only report partial results.

Whenever models were publicly available, we re-ran tests using published weights, rather than citing results from published papers; therefore, our results sometimes differ from the published results. For instance, in the case of CLIP-ViT-L-14, we found that certain tasks varied by as much as $6\%$ from the published figures on OpenAI's website. These are outliers, and generally our metrics match the aforementioned published results.

\section{Results and Discussion}
\label{sec:results_remarks}

\begin{table*}[t]
  \caption{\textbf{OOD-robust models share key properties.} \sl Our core findings on robustness are divided into artificial and natural examples. We order our results by accuracy on the main task, in descending order. Three models were robust to natural distribution shifts -- all of them used contrastive VL pretraining on massive datasets but varied in most other respects, including dataset used, image backbone, and hyperparameters. No model was robust to artificial distribution shifts. Robust models are in $\textbf{bold}$. Key: Val denotes validation accuracy on ImageNet-2012, NR denotes the average natural OOD-robustness measured on our testbed datasets, AR denotes the artificial OOD-robustness measured on our testbed datasets, VL denotes vision-language pretraining.}
  \label{tab:main_results}
  \vskip 0.15in
  \centering
  \begin{tabular}{c l@{\kern1em}l@{\kern1em}l@{\kern1em}l@{\kern1em}l@{\kern1em}l@{\kern1em}c@{}}
    \toprule
    Acc. Regimes & Model & Val & NR & AR & VL & Data & Backbone\\
    \midrule
    \multirow{8}{*}{High} & EfficientNet-L2-NS & 88.32 & 80.71 & 81.88 & \red{\xmark} & \yellow{300M} & EfficientNet \\
    {} & FixResNext101-V2 & 86.36 & 79.48 & 65.59 & \red{\xmark} & \green{1B} & ResNet \\
    {} & SEER & 85.8 & 67.79 & N/A & \red{\xmark} & \green{1B} & RegNet \\
    {} & BASIC & 85.7 & $\textbf{97.29}$ & N/A & \green{\cmark} & \green{6.6B} & ViT \\
    {} & ViT-L-16-384 & 85.15 & 66.17 & 70.86 & \red{\xmark} & \yellow{300M} & ViT \\
    {} & RN-152-SimCLR-FT & 81.04 & 57.63 & 62.8 & \red{\xmark} & \red{$\le$300M} & ResNet\\
    {} & ALIGN & 76.4 & $\textbf{98.19}$ & N/A & \green{\cmark} & \green{1.8B} & EfficientNet \\
    {} & CLIP-ViT-L-14 & 75.53 & $\textbf{93.83}$ & 67.6 & \green{\cmark} & \green{400M} & ViT \\
    \midrule
    \multirow{4}{*}{Medium} &  Resnet-34 & 73.31 & 42.26 & 47.13 & \red{\xmark} & \red{$\le$300M} & ResNet-34\\
    {} & CLIP-ViT-B-32 & 63.3 & 76.7 & 56.68 & \green{\cmark} & \green{400M} & ViT\\
    {} & OpenCLIP-ViT-B-32-L400M & 60.3 & 76.33 & 53.33 & \green{\cmark} & \green{400M} & ViT\\
    {} & CLIP-RN50 & 59.8 & 70.72 & 39.91 & \green{\cmark} & \green{400M} & ResNet \\
    {} & AlexNet & 56.52 & 32.39 & 29.9 & \red{\xmark} & \red{$\le$300M} & CNN\\
    \midrule
    \multirow{5}{*}{Low} & SLIP-ViT-L-YFCC15M & 47.9 & 52.34 & 54.74 & \green{\cmark} & \red{$\le$300M} & ViT\\
    {} & RN50-Adv-Smooth & 44.26 & 38.75 & 64.69 & \red{\xmark} & \red{$\le$300M} & ResNet\\
    {} & SLIP-ViT-B-16-CC12M & 40.7 & 59.28 & 44.4 & \green{\cmark} & \red{$\le$300M} & ViT\\
    & RN-Subsample & 36.7 & 27.25 & 25.86 & \red{\xmark} & \red{$\le$300M} & ResNet \\
    & OpenCLIP-RN50-CC12M & 35.98 & 71.52 & 29.85 & \green{\cmark} & \red{$\le$300M} & ResNet \\
    & OpenCLIP-RN101-YFCC15M & 34.93 & 59.03 & 31.86 & \green{\cmark} & \red{$\le$300M} & ResNet \\
    \bottomrule
  \end{tabular}
\end{table*}

Table~\ref{tab:acc_regimes} provides details of our meta-analysis. We observed several commonalities among models that are OOD-robust, which we highlight as follows.

\subsection{Vision-Language Contrastive Pretraining}
\label{subsec:vlconst}
Our findings corroborate those of~\cite{Radford2021LearningTV}: vision-language models are capable of achieving extremely high natural robustness (above $90\%$), and in this respect, they appear to differ substantially from other training objectives ~\cite{Taori2020MeasuringRT}. The effect is pronounced and reliable, originating in the low-accuracy regime and persisting throughout training. (See Table~\ref{tab:acc_regimes}).

Some of the examples of this phenomenon are truly striking. For instance, our experiments on OpenCLIP reveal that, after only 30 epochs of training, OpenCLIP, trained on 8 GPUs, exhibits proportionate out-of-distribution robustness within \emph{ten} percent of the fully-trained SEER model~\cite{Goyal2022VisionMA}, with $\sim10B$ parameters and trained with a SwAV~\cite{caron2020unsupervised} objective on almost 500 GPUs.

\subsection{Artificial Distribution Shifts}
\label{subsec:artif_dist_shift}

Reported results on robustness to artificial distribution shifts are somewhat scarcer than those on natural distribution shifts. To the best of our knowledge, none of the large-scale private models have reported results on ImageNet-c or Stylized-ImageNet.

However, our own experiments (Table 3 in the appendix) seem to indicate that artificial distribution robustness may be a much bigger challenge than natural distribution robustness, since none of our models were OOD-robust to artificial distribution shifts. This may have to do with the methodologies used to generate these particular datasets, which include many samples that have proven extremely challenging even for humans to classify.~\cite{geirhos2018imagenettrained}

Our experiments also affirm that vision-language pretraining does not reliably make a model more robust to artificial distribution shifts as it appears to do with natural distribution shifts. This, too, follows the scaling laws demonstrated by ~\cite{Taori2020MeasuringRT} and ~\cite{Miller2021AccuracyOT}.

\subsection{Model Architecture and Dataset Size}
\label{subsec:obs_model_arch}

The results reported in the CLIP and ALIGN papers demonstrate that ViT-, ResNet- and EfficientNet backbones can be used to train OOD-robust models.

Model capacity is crucial. CLIP experiments show us that lower-capacity models such as ResNet-50 and ViT-B-32 fail to improve after the model is saturated. The training curves of most of the larger vision models suggest that it is very straightforward to know when such a limit has been reached ~\cite{Radford2021LearningTV}. Precisely how large a model architecture is necessary remains an open question, but the results shown in ~\cite{Radford2021LearningTV} strongly suggest that even larger models than ViT-L may prove beneficial.

A precise cutoff for the number of data samples required to train an OOD-robust model remains elusive. The effect of dataset size on robustness of non-VL models, which are more widely available, show robustness benefits for models trained with at least 300M samples, but shows few marginal gains to robustness above that threshold.

\subsection{Other Parameters}

Of the remaining covariates, the question of batch size is a matter of practical concern. Most VL-models examined in our meta-review used a very large batch size, including all models (of which we are aware) that belong to the high-accuracy regime. 

Our experiments with OpenCLIP (batch size 128) and SLIP (batch size 4096), the best comparison pair available to us, show comparable OOD robustness for both models. However, it is likely that neither model was trained to the point of saturation, and it is possible that the model may learn harder examples more effectively later in the training process.~\cite{Tu2020AnES}

\subsection{Common Properties of Robust Models}
\label{subsec:model_archs}

While our meta-review is likely not comprehensive, we observe that the following qualities consistently recur:
\begin{enumerate}[noitemsep, parsep=0pt]
    \item Vision-language contrastive pretraining objective.
    \item $\ge$ 400m training samples with noisy labels.
    \item A high-capacity ViT or ResNet backbone, such as ViT/L or EfficientNet.
    \item Sufficiently many training epochs to reach the high accuracy regime
\end{enumerate}
Our results show that among more than two hundred models, many evaluated for the first time in our meta-review, no model achieved robustness without these properties. Further, we have not encountered any model which fulfilled these requirements and failed to achieve robustness.

Current best practices, therefore, may involve focusing on these four properties as prerequisites. We further recommend carefully probing the effect of batch size $>$ 4096 and dataset sizes between $100$M and $400$M when designing future OOD-robust classifiers, as the effects of these may be present but are not as consistently observed.

\section{Conclusions}
\label{sec:conclusions}

In this meta-review we identify several commonalities among distributionally-robust models. This, combined with results from other recent studies, enables us to suggest best practices to achieve OOD-robustness of models that are trained up to the high accuracy regime. 

We also make qualitative remark on key outstanding questions, such as the precise role batch size and internet-scale datasets play in training robust models. 

Overall, we believe that continued attention to these questions will pave the way towards achieving even superior models than those that are currently available. 


\bibliography{commonalities_icml}
\bibliographystyle{icml2022}

\newpage
\appendix
\onecolumn
\section{Appendix}

\subsection{Experimental Design}
\label{sec:experimental_design}

Because many potential points of comparison exist in a large-scale study, our methodology focuses on a relatively narrow set of variables. For this reason, our study largely follows an existing methodology of Taori \emph{et al.} and Miller \emph{et al.}~\cite{Taori2020MeasuringRT,Miller2021AccuracyOT}. In these papers, over two hundred models are examined under very similar conditions, including CNNs, visual transformers, self-supervised models, logistic regression, nearest neighbors, and kernel machines, making it an ideal starting point for investigating questions of robustness.

We select for our experiments an extended subset of the models examined in the above papers, extending it in three important ways. First, we evaluate the robustness of several new models which were not tested in the paper. Second, we present an expanded set of artificial distribution shift results for models which were in the original study, but for which artificial distribution robustness results were previously unavailable. Third, we collect and compare robustness results which have been published, with a special attention to contrastive models.

A complete list of the representative sample of models considered in this paper, along with a breakdown of the key properties we compare, can be seen in \ref{tab:main_results}.

The pretrained SLIP weights used in this paper can be can be accessed via \href{https://github.com/facebookresearch/SLIP}{their repository}. The OpenCLIP weights and the framework used to train them can be found in ~\cite{ilharco_gabriel_2021_5143773}.

The provenance of the remaining models, including links to pretrained weights, can be referenced in ~\cite{Taori2020MeasuringRT}.

We fully acknowledge that opinions on what constitutes a 'representative sample' of models may differ, and we encourage the curious reader to explore the interactive testbed in ~\cite{Taori2020MeasuringRT} and ~\cite{Miller2021AccuracyOT}, where hundreds of additional model results can be examined. Furthermore, since both the models and the testbed are publicly available, the reader is also free to evaluate and compare additional models beyond what we present here.

Wherever possible, we use the ImageNet-testbed benchmark suite from that same paper, in order to provide a consistent grounds for comparison between them models. Where this is not possible, we run inference using the same label set as provided in that benchmark.

We quantify robustness to distribution shift as the ratio of average OOD-robustness to in-distribution robustness:

\begin{equation*}
  \frac{\text{avg}(OOD)}{ID}
  \label{eq:ood_id}
\end{equation*}

For more information on the datasets tested here, please refer to \ref{sec:dataset_details}.

Wherever possible, we consider classifier accuracy as a relative, rather than absolute, measurement -- we are interested in consistency above all. However, we also provide raw accuracy scores in the appendix for reference.

\subsection{Additional detail on datasets}
\label{sec:dataset_details}

Our selection of natural distribution datasets is \textsc{ImageNet-a}, ~\cite{hendrycks2021nae} \textsc{ImageNet-r}, ~\cite{hendrycks2021many}	\textsc{ImageNet-s}, ~\cite{Wang2019LearningRG} \textsc{ImageNet-V2}, ~\cite{Recht2019DoIC} and \textsc{ObjectNet} ~\cite{Borji2020ObjectNetDR}. We use the matched-frequency-format-val version of ImageNet-V2, and the 1.0-beta version of ObjectNet. We also provide a small sample of our comparison datasets for randomly selected classes in ImageNet.

\begin{figure*}
  \centering
  \includegraphics[width=0.95\linewidth]
                  {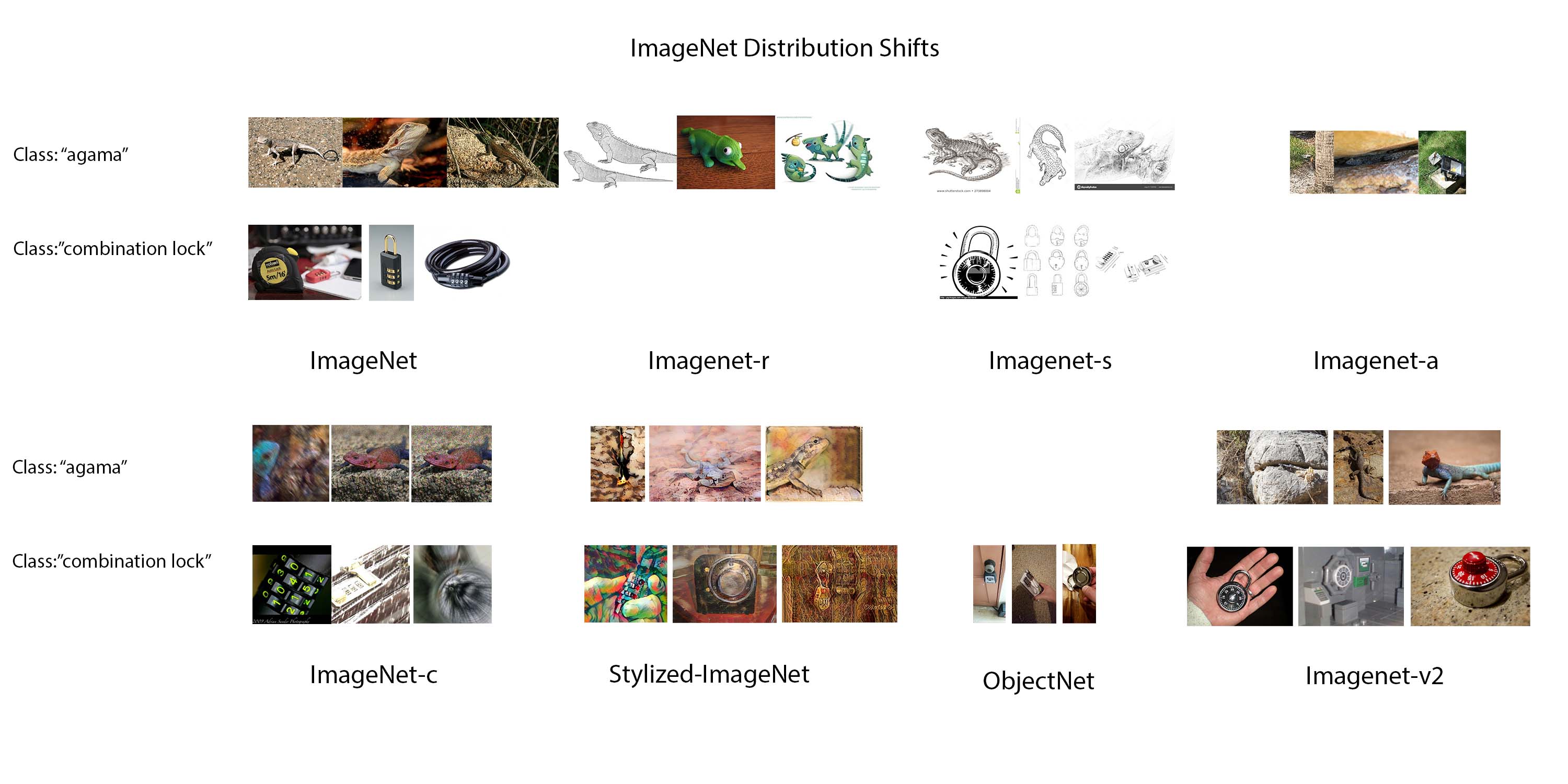}
  \caption{\textbf{Examples from Datasets used for our Robustness Tests.}\sl Two classes from the ImageNet validation set (upper left) and their distribution-shifted counterparts. Blank spaces indicate missing classes. Artificial distribution shifts show the model perturbations of images it has already classified, whereas natural distribution shifts show the model un-perturbed but novel images.}
  \label{fig:dataset_examples}
\end{figure*}

We select \textsc{ImageNet-C-in-memory-avg}~\cite{Hendrycks2019BenchmarkingNN}, \textsc{ImageNet-C-on-disk-avg}, and \textsc{Stylized-ImageNet}~\cite{geirhos2018imagenettrained} as test-beds for our artificial distribution robustness experiments. ImageNet-C is a set of common visual corruptions (blur, noise) applied to the ImageNet test-set.  The former is computed by the GPU at test time, the latter evaluates a fixed set of JPEGs. Our experiments show that 10-percent ImageNet-C subset results are very good approximations of the full dataset -- we use them on all models except for the SLIP models, which are incompatible with CLIP's framework, and therefore with the testbed. Following ~\cite{Taori2020MeasuringRT, Miller2021AccuracyOT}, for all models except SLIP, we evaluate ImageNet-C both in-memory and on-disk, sometimes finding significant discrepancies between the two. Our averaging method treats all ImageNet-C in-memory corruptions as a single dataset, and all ImageNet-C on-disk corruptions as another single dataset.

Stylized-ImageNet was created by applying AdaIN style transfer to ImageNet images. These generated images range from highly recognizable to extremely abstract. It is the most challenging dataset we evaluate in this study, with the best classifier achieving a little over $55\%$ accuracy.

We observe that there are important differences in the two distribution shifts that may partly account for our findings on them; artificial distribution shifts show the model perturbations of images it has already classified, whereas natural distribution shifts show the model un-perturbed but novel images.

\subsection{OOD robustness scales disproportionately with in-distribution robustness}
\label{subsec:ood_indist}
\begin{table*}[b]
  \caption{\textbf{Robustness trends emerge in the low-accuracy regime.} \sl Comparing VL and non-VL models in each accuracy regime, we can see that on average, OOD-robustness trends persist as the model is trained to higher accuracy. We also see a stark contrast in natural and artificial distribution robustness -- under artificial shifts, VL models perform no better than non-VL models.}
  \label{tab:acc_regimes}
  \vskip 0.15in
  \centering
  \begin{tabular}{l@{\kern1em}l@{\kern1em}l@{\kern1em}lc@{}}
    \toprule
    Model Class & Acc Regime & Nat. Range & Artif. Range \\
    \midrule
    Non-VL & Low & 27-39 & 26-65 \\
    VL & Low & 52-72 & 30-51 \\
    Non-VL & Medium & 32-42 & 30-47\\
    VL & Medium & 71-77 & 40-57 \\
    Non-VL & High & 57-81 & 63-79 \\
    VL & High & 95-97 & 68 \\
    \bottomrule
  \end{tabular}
\end{table*}

Recent studies have suggested that models in the low and medium accuracy regimes appear to follow the scaling laws observed by Taori et al~\cite{Taori2020MeasuringRT} and Miller et al~\cite{Miller2021AccuracyOT}. Our experiments confirm these observations on a wider range of vision-language models and over a wider range of datasets. This is important, given that one of the aims of our work is to better predict the robustness of unseen classifiers. The above-cited scaling law estimates allow us to make a better guess at an early checkpoint whether a model has the potential to achieve robustness.

Our experiments confirm that models which are eventually trained to robustness perform significantly better on the metric in the low and medium accuracy regimes. We also note that natural OOD-robustness scales reliably with model accuracy in vision-language models, and that it occupies a narrower range, compared to non-vision-language classifiers. See Table~\ref{tab:main_results} for detailed results.

This appearance of uniformity, however, conceals another interesting observation -- in the low and medium accuracy regimes, vision-language models are more robust on certain datasets (ImageNet-V2, ImageNet-r) than others (ImageNet-a, ImageNet-s). This effect is less pronounced in non-VL models, and may be caused by informative contrastive counterexamples the model has yet to encounter.~\cite{Tu2020AnES}

\subsection{Noisy Student Training}
\label{subsec:nst}

One near-exception to the domination of vision-language contrastive models can be found in the area of knowledge distillation. Noisy Student Training~\cite{DBLP:journals/corr/abs-1911-04252} extends the idea of self-training and distillation with the use of equal-or-larger student models and noise added to the student during learning. An EfficientNet teacher generates pseudo-labels on 300M unlabeled images. A larger EfficientNet student model then trains on the combination of labeled and pseudo labeled images.

The effects on robustness are similar, but less pronounced and less consistent, to those seen in VL training. Specifically, this approach is the most robust, both in proportionate and in absolute terms, at handling artificial distribution shifts.

In fact, there appears to be a curious counterbalance between the two training approaches, as can be seen in Table 3. For all of the tasks we evaluated, either CLIP ViT-L or Noisy Student were the most robust, but with the exception of ImageNet-a, there is a marked performance discrepancy between the models on every single task.

Although speculative explanations immediately suggest themselves, we leave the explanation for this curious inverse relationship to future work.

We do note that data augmentation may have helped Noisy Student's robustness to artificial distribution shifts, and it is possible that a VL model trained in a similar manner to CLIP, but with data augmentation added to the training procedure, would show improvements in this area.

\begin{table*}[b]
    \caption{\textbf{Raw accuracy scores on the datasets.} \sl The above table collects the raw accuracy scores for each model on each dataset in the testbed, including only the models we were able to test directly. We also report the proportional 95\% confidence interval average for all models on each dataset, rounded up to the nearest .005. CLIP and Noisy Student, the most OOD-robust VL and non-VL models we tested directly, exhibit largely complementary behavior on per-task accuracy comparison. CLIP performs better on natural distribution shifts, while Noisy Student performs better on artificial ones. COLUMNS FROM LEFT TO RIGHT: ImageNet-2012, ImageNet-a, ImageNet-r, ImageNet-sketch, ImageNet-V2, ObjectNet, ImageNet-C (in memory, 10\% subset), ImageNet-C (on disk, 10\% subset), Stylized ImageNet}
   \label{tab:acc_vl_nonvl}
   \vskip 0.15in
  \centering
  \begin{tabular}{l@{\kern1em}l@{\kern1em}l@{\kern1em}l@{\kern1em}l@{\kern1em}l@{\kern1em}l@{\kern1em}l@{\kern1em}l@{\kern1em}lc@{}}
    \toprule
    Model & IN-VAL & IN-A & IN-R & IN-S & IN-V2 & ObjNet & IN-c-mem & IN-c-disk & Style-IN \\
    EfficientNet-L2-NS & 88.32 & 84.85 & 74.67 & 47.64 & 80.85 & 68.45 & 83.92 & 76.05 & 56.99 \\
    FixResNext101-V2 & 86.36 & 68.41 & 79.9 & 59.14 & 77.94 & 57.85 & 72.63 & 59.63 & 37.66 \\
    ViT-L-16-384 & 85.15 & 53.85 & 54.75 & 43.28 & 75.37 & 54.49 & 75.27 & 65.74 & 40.03 \\
    RN-152-SimCLR-FT & 81.04 & 33.85 & 47.3 & 35.06 & 70.35 & 46.94 & 63.8 & 57.42 & 31.49 \\
    CLIP-ViT-L-14 & 76.2 & 71.15 & 87.74 & 59.57 & 75.87 & 67.7 & 60.98 & 58.1 & 35.46 \\
    ResNet-34 & 73.31 & 1.87 & 36.07 & 23.47 & 60.87 & 32.62 & 42.37 & 39.6 & 21.69 \\
    CLIP-ViT-B-32 & 63.3 & 31.38 & 69.25 & 42.37 & 55.77 & 43.96 & 44.26 & 42.13 & 21.25 \\
    OpenCLIP-ViT-B-32-L400M & 60.3 & 19.63 & 70.73 & 46.42 & 52.2 & 41.15 & 39.45 & 36.97 & 20.08 \\
    CLIP-RN50 & 59.8 & 22.8 & 60.46 & 35.38 & 52.53 & 40.28 & 31.48 & 27.43 & 12.71 \\
    Alexnet & 56.52 & 1.77 & 21.77 & 10.71 & 43.43 & 13.86 & 23.1 & 20.18 & 7.41 \\
    SLIP-ViT-L-YFCC15M & 47.9 & 23.17 & 33.05 & 19.01 & 42.09 & 24.05 & 33.41 & 31.88 & 14.9 \\
    RN50-Adv-Smooth & 44.26 & 2.06 & 27.07 & 11.53 & 33.23 & 11.86 & 39.45 & 35.91 & 10.54 \\
    SLIP-ViT-B-16-CC12M & 40.7 & 5.5 & 37.68 & 29.2 & 34.58 & 20.17 & 21.91 & 21.76 & 10.7 \\
    RN-Subsample & 36.7 & 1.41 & 11.32 & 3.17 & 26.94 & 7.17 & 12.75 & 12.72 & 2.98 \\
    OpenCLIP-RN50-CC12M & 35.98 & 7.57 & 44.64 & 23.46 & 30.44 & 22.55 & 13.54 & 11.56 & 7.11 \\
    OpenCLIP-RN101-YFCC15M & 34.93 & 15.68 & 26.28 & 8.89 & 33.52 & 18.73 & 14.63 & 12.91 & 5.85 \\
    U.B. Confidence Intvl (+/-) & 0.005 & 0.015 & 0.005 & 0.005 & 0.01 & 0.005 & 0.005 & 0.005 & 0.005 \\
    \midrule
    \bottomrule
  \end{tabular}
\end{table*}

\end{document}